\documentclass{article}
\usepackage{ijcai19}
\usepackage{epsfig}
\usepackage{times}
\usepackage{soul}
\usepackage{url}
\usepackage[hidelinks]{hyperref}
\usepackage[utf8]{inputenc}
\usepackage[small]{caption}
\usepackage{graphicx}
\usepackage{amsmath}
\usepackage{amssymb}
\usepackage{booktabs}
\usepackage{algorithm}
\usepackage{algorithmic}
\usepackage{color}
\usepackage{bm}
\title{Image Super-Resolution Using Attention Based DenseNet with Residual Deconvolution}

\author{
Zhuangzi Li$^1$ \\
$^1$Beijing Technology and Business University \\
lizhuangzii@163.com
}

\begin{document}

\maketitle

\begin{abstract}
Image super-resolution is a challenging task and has attracted increasing attention in research and industrial communities. In this paper, we propose a novel end-to-end \underline{A}ttention-based \underline{D}enseNet with \underline{R}esidual \underline{D}econvolution named as ADRD. In our ADRD, a weighted dense block, in which the current layer receives weighted features from all previous levels, is proposed to capture valuable features rely in dense layers adaptively. And a novel spatial attention module is presented to generate a group of attentive maps for emphasizing informative regions. In addition, we design an innovative strategy to upsample residual information via the deconvolution layer, so that the high-frequency details can be accurately upsampled. Extensive experiments conducted on publicly available datasets demonstrate the promising performance of the proposed ADRD against the state-of-the-arts, both quantitatively and qualitatively. 
\end{abstract}

\section{Introduction}

Image super-resolution aims at recovering high-resolution (HR) images from it's low-resolution (LR) versions. By far, it has been widely applied to various intelligent image processing applications, e.g. license plate recognition \cite{DBLP:conf/mm/LiuLMC17}, video surveillance \cite{DBLP:journals/tip/ZouY12}. However, image super-resolution is an inherently ill-posed problem since the mapping from the LR to HR space can have multiple solutions. To deal with this issue, various promising super-resolution approaches have been proposed in the past years \cite{DBLP:journals/pami/KimK10,DBLP:conf/cvpr/YangLC13,DBLP:journals/tog/FreedmanF11,DBLP:conf/cvpr/TaiY017,Hui_2018_CVPR}.

In image super-resolution, recovering high-frequency is a key problem that the super-resolved images should be full of edges, textures, and other details. Recently, convolutional neural networks (CNNs) are gradually applied to image super-resolution relying on CNN's great approximating to capture high-frequency.
Dong et al. firstly introduced CNN's architecture for the image super-resolution in \cite{DBLP:journals/pami/DongLHT16}. Later, a series of CNNs \cite{DBLP:conf/cvpr/KimLL16a,DBLP:conf/cvpr/KimLL16,DBLP:conf/cvpr/TaiY017,DBLP:conf/cvpr/LaiHA017,RDN} try to solve the problem by increasing network depth. Shortcut connections \cite{DBLP:conf/cvpr/KimLL16,DBLP:conf/cvpr/TaiY017,DBLP:conf/cvpr/LaiHA017,RDN} demonstrates the power of recovering high-quality images. As a kind of shortcut connections, dense connections are introduced in \cite{DBLP:conf/iccv/0001LLG17,SRDDNet} to recover images by extracting additional information from hierarchical features. However, the above methods treat all hierarchical features equally and lack of the flexibility to select valuable features. Moreover, spatial features are not well explored, resulting in the loss of high-frequency information during feedforward.
\begin{figure}[t]
\centering

\begin{minipage}[t]{0.23\linewidth}
\centering
\includegraphics[width=1\linewidth]{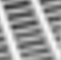}
\small{Bicubic}
\end{minipage}
\begin{minipage}[t]{0.23\linewidth}
\centering
\includegraphics[width=1\linewidth]{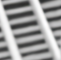}
\small{RDN}
\end{minipage}
\begin{minipage}[t]{0.23\linewidth}
\centering
\includegraphics[width=1\linewidth]{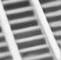}
\small{Ours}
\end{minipage}
\begin{minipage}[t]{0.23\linewidth}
\centering
\includegraphics[width=1\linewidth]{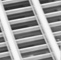}
\small{HR}
\end{minipage}
\caption{\label{fig:firstExample} Side-by-side image super-resolution comparisons of bicubic interpolation, the state-of-the-art RDN, our method and ground-truth HR image.}
\end{figure}
%However, they treat all previous hierarchical features equally, so the network is lack of informative clues during reconstruction. Though various feature maps can be obtained by dense connections, the internal information is not further explored. 
Furthermore, high-frequency information can not be well upscaled by the conventional deconvolution as stated in \cite{DBLP:conf/eccv/DongLT16,DBLP:journals/corr/MaoSY16a,DBLP:conf/iccv/0001LLG17}.

 \begin{figure*}[t]
  \centering
  \mbox{} \hfill
   \hfill
  \includegraphics[width=0.92\linewidth]{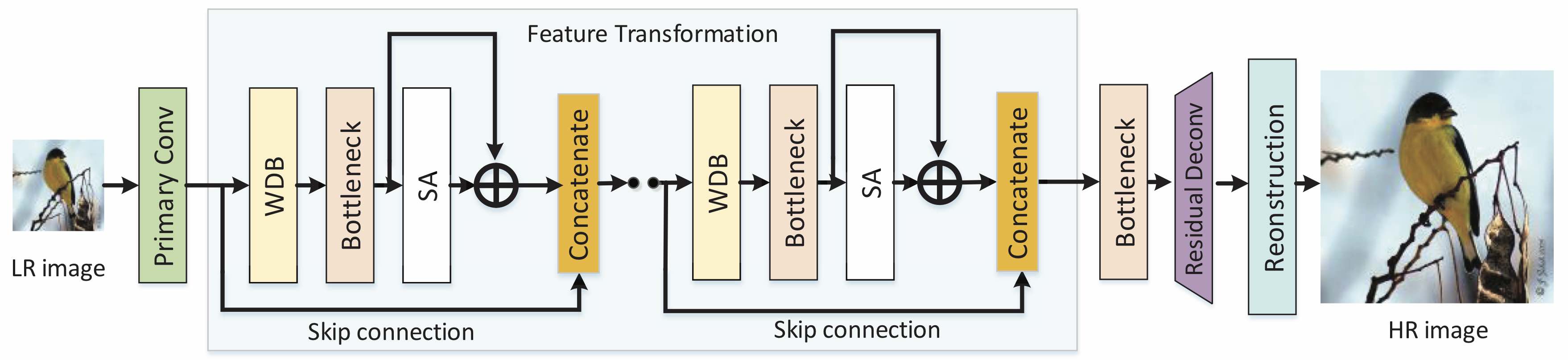}
  \hfill \mbox{}
  \caption{\label{framework}%
           Framework of our attention based DenseNet with Residual Deconvolution (ADRD) for image super-resolution.}
\end{figure*} 
To practically tackle the above-mentioned problems, we propose a novel image super-resolution framework based on attention based densely connected network (DenseNet) with a residual deconvolution (ADRD). As shown in Figure \ref{fig:firstExample}, our method can generate high-quality super-resolved images compared with the state-of-the-art RDN \cite{RDN}. Specifically, weighted dense block (WDB) is proposed, where features from preceding layers are weighted into current layers. In such a way, different hierarchical features can be effectively combined by its significance. Then, we present a novel spatial attention module by learning feature residual from the WDB, enhancing the informative details for feature modeling, thus high-frequency regions can be highlighted. Further, an innovative upsampling strategy is devised that allows abundant low-frequency information to be bypassed through interpolation and focus on accurately upsampling high-frequency information.
To summarize, the main contributions of this paper are three-fold:
\begin{itemize}
\item We propose ADRD for image super-resolution and achieve state-of-the-art performance.
\item  Proposing a weighted dense block to adaptively combine valuable features.
\item  Presenting a spatial attention method to emphasize high-frequency information.
\item  An innovative residual deconvolution algorithm is proposed for upsampling.
\end{itemize}

Our anonymous training and testing codes, final model, and supplementary experimental results are available at website: {\color[RGB]{243,100,136}\textsl{https://github.com/IJCAI19-ADRD/ADRD}}.

\section{Our method}
\label{our method}

The framework of ADRD is shown in Figure \ref{framework}, which contains 4 parts. The LR image is firstly fed into a $3\times3$ convolution layer and PReLU \cite{PReLU} to get primary feature maps. Then, the primary feature maps are put into $4$-groups based feature transformation.

In each group, weighted dense block (WDB) can obtain deeply diversified representations by weighted dense connections. A bottleneck layer would compress increasing feature maps extracted from the WDB. Next, the spatial attention (SA) module receives the compressed features and generate a residual output by attentive maps. The residual output integrates with the compressed features, thus enhanced features are obtained. For easy training and increasing the width of network, skip connections \cite{DBLP:conf/iccv/0001LLG17,SRDDNet} are introduced to make the input feature maps of the WDB concatenate the enhanced features. In the end of feature transformation, a bottleneck layer works for compressing global features.

The transformed features are upsampled by a residual deconvolution approach which amplifies feature maps to HR's sizes. Finally, the reconstruction component, a 3-channel output convolution layer, reconstructs feature maps to the RGB channel space, and the prospective HR image can be obtained. Our contributions, weighted dense block, spatial attention module, as well as the residual deconvolution strategy, will be illustrated in next sections in detail.

\subsection{Weighted dense block}
Dense connections can alleviate the vanishing-gradient problem, strengthen feature propagation and substantially reduce the number of parameters \cite{DBLP:conf/cvpr/HuangLMW17}. Inspired by \cite{SRDDNet}, we take advantage of dense connection for capturing diverse information from different hierarchies. In dense blocks of dense connection network, dense layers are sequentially stacked, and have short paths from previous dense layers. Consequently, the $\ell$-th dense layer receives the feature-maps of all preceding layers. Let's $x_{0},...,x_{\ell-1}$ denote the input feature maps of the $\ell$-th dense layer. Then the output of $x_{l}$ can be formulated as:
\begin{equation}
x_{\ell} = H_{\ell}([x_{0},x_{1},...,x_{\ell-1}]), \label{eq2}
\end{equation}
where $[x_{0},x_{1},...,x_{\ell-1}]$ denotes channel-wise concatenation of feature maps. $H_{l}$ denotes as composite function which consists of Rectified Linear Units (ReLUs), a $1 \times 1$ convolution layer and a $3 \times 3$ convolution layer. A group of dense layers are combined as a dense block. However, in existing dense block based methods \cite{SRDDNet,RDN}, they treat previous level features equally. Consequently, some beneficial features cannot be well represented, and some vulgar features will restrain the final super-resolution performance.

% \begin{figure}[h]
%  \centering
%  \mbox{} \hfill
%   \hfill
%  \includegraphics[width=1.0\linewidth]{WDB}
%  \hfill \mbox{}
%  \caption{\label{WDB}%
%           The structure of  (WDB).}
%\end{figure} 
\begin{figure}[h]
\centering
\includegraphics[width=0.9\linewidth]{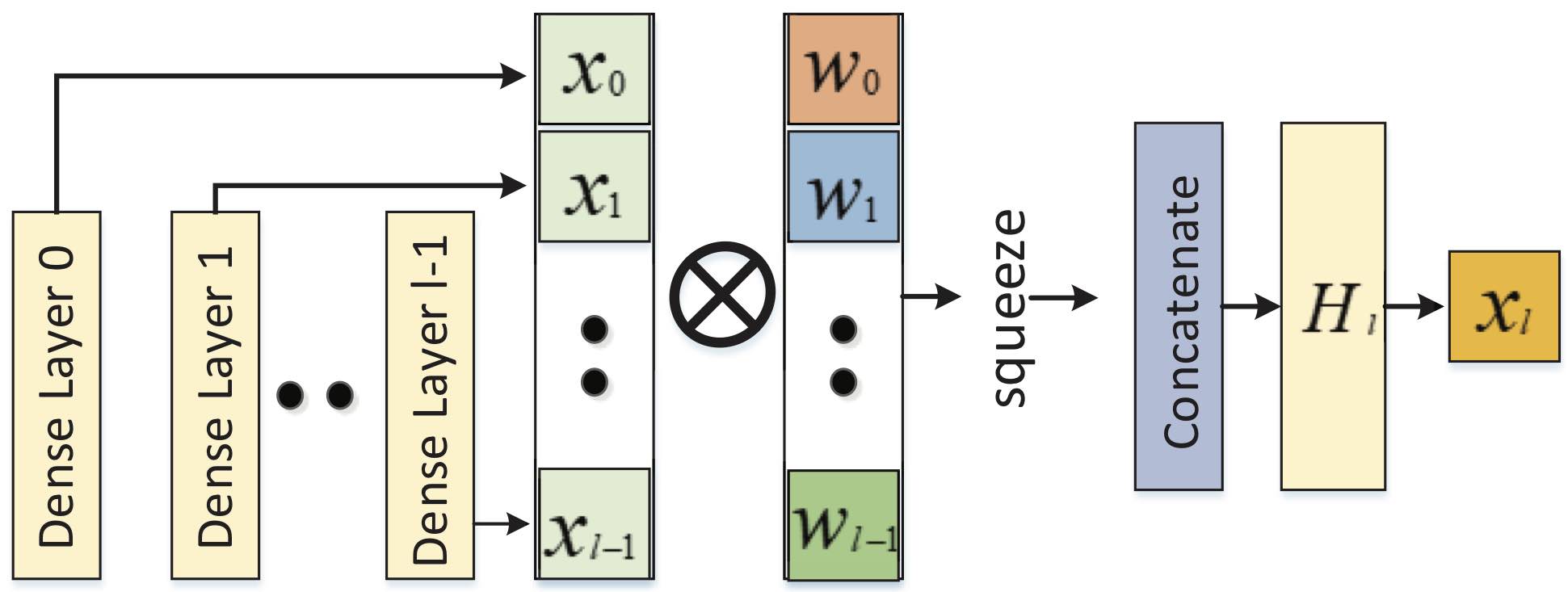}
\caption{\label{differentdense}%
           Calculation of WDB in the $l$-th dense layer. $\otimes$ denotes element-wise product. }
\end{figure} 
To solve the problem, we propose WDB. It aims to increase the flexibility during feature combinations by adaptively learning a group of weights. As shown in Figure \ref{differentdense}, each dense layer assigns a set of weights to the preceding layers. Thus, valuable features will be adequately explored in the current level, while restrain unimportant features will be suppressed.. The WDB output of the $l$-th layer can be formulated as:
\begin{equation}
x_{\ell} = H_{\ell}([\omega_{0} \cdot x_{0}, \omega_{1} \cdot x_{1},..., \omega_{\ell-1} \cdot x_{\ell-1}]), \label{eq3}
\end{equation}
where $\omega$ is the weight of preceding level features. From Eq. \ref{eq2} and Eq. \ref{eq3}, we can see that dense connection is a special case of the weighted dense connections in the condition of $\omega =1$. Notably, the channel number of $x$ is called as growth rate $G$, which is equal in all block.
%Notably, the WDB design is not only carried out in the image super-resolution task, but also image inpainting, and image denoising. The $1 \times 1$ convolutional kernel can be introduced as bottleneck layer before each $3 \times 3$ convolution to reduce the number of input feature maps, and improve computational efficiency.

\subsection{Spatial attention}
\begin{figure}[h]
\centering
\includegraphics[width=1.0\linewidth]{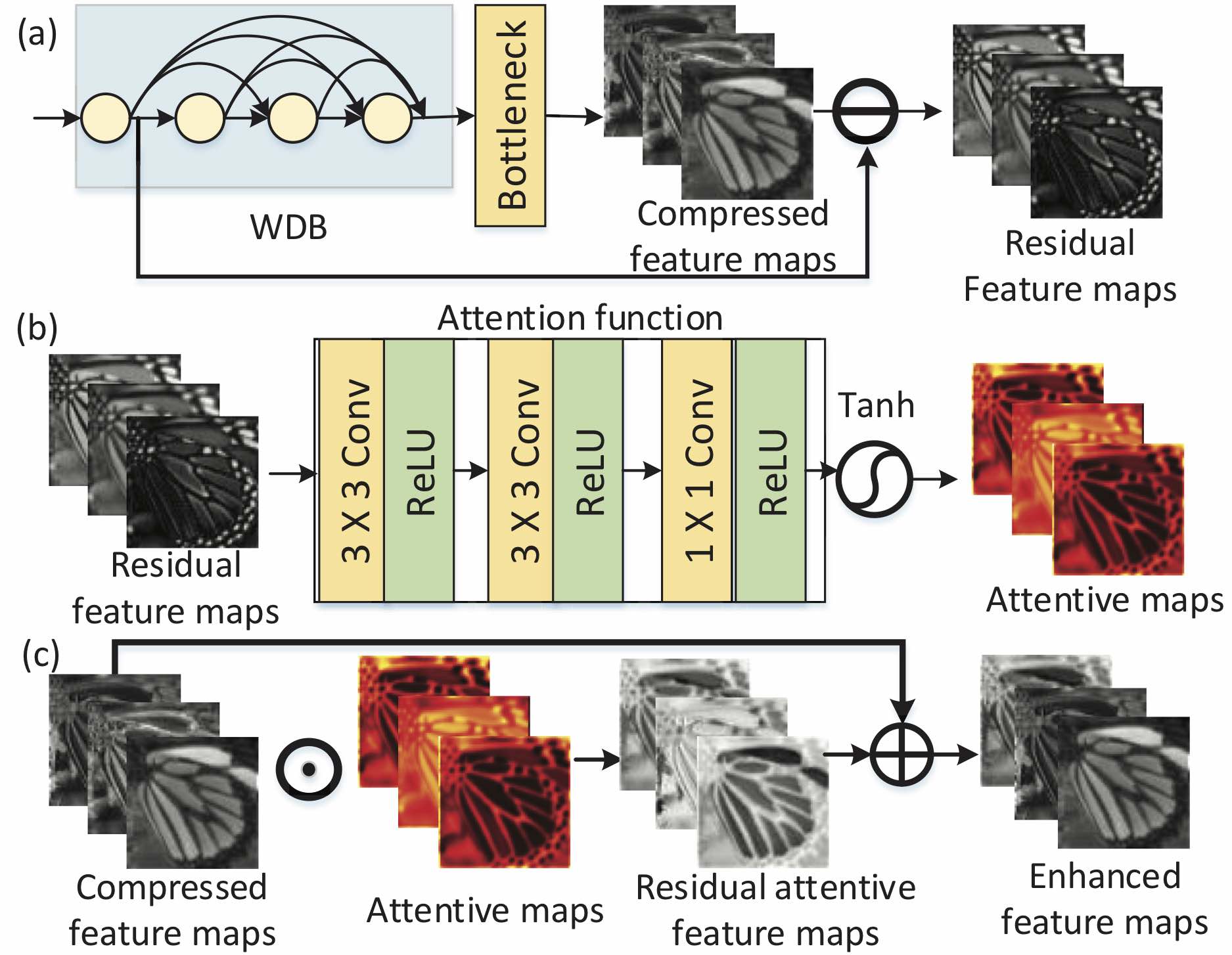}
\caption{\label{FSA}%
         Flowchart of spatial attention: (a) Residual features generation. (b) Attentive maps generation. (c) Enhanced feature maps generation.}
\end{figure}

The spatial attention module aims to enhance the high-frequency information by learning a group of attentive maps. The attentive maps can give large weights for informative regions.
Flowchart of spatial attention is shown in Figure \ref{FSA}.

Detailedly, the spatial attention module includes three stages: (a) Residual features generation. (b) Attentive maps generation and (c) Enhanced feature maps generation. In step (a), the information residual between the head layer in the WDB (denoted as $X_{in}$) and the features compressed from bottleneck layer (denoted as $X_{bot}$) is computed. The bottleneck in here is composed by a $1 \times 1$ convolutional layer and a ReLU function, which guarantees $X_{bot}$ should have the same channel number as $X_{in}$.
The residual feature maps $X_{res}$ can be obtained as:
\begin{equation}
X_{res} = |X_{in} - X_{bot}|.
\end{equation}
In step (b), the residual feature maps are then fed into an attention function $f_{att}$, which contains two $3\times3$ convolutional layers, and a $1 \times 1$ convolutional layer. Thus, attentive maps are generated and formulated as:
\begin{equation}
X_{att} = \mathrm{Tanh}(f_{att}(X_{res})), 
\end{equation}
where $Tanh$ represents the tangent function, which has larger gradients than Sigmoid near to 0.
In step (c), $X_{att}$ and $X_{bot}$ are combined to generate residual attentive features $X_{ram}$:
\begin{equation}
X_{ram} = X_{att} \odot X_{bot},
\end{equation}
where $\odot$ is Hadamard product. Based on the residual attentive feature maps and the $X_{bot}$, the enhanced feature maps are then generated by:
\begin{equation}
X_{enhanced} = \lambda X_{ram} + X_{bot},
\end{equation}
%$X_{enhanced}$ concatenates the input features of WDB by skip connections, then puts into to the next WDB.
where $\lambda$ is a hyper-parameter to keep an attention level. Our attention method can extract the content information of features, and learn to generate attentive maps. The super-resolved images tend to be clearer and sharper, because $X_{enhanced}$ contains more high-frequency information.

\subsection{Residual deconvolution}
Deconvolution is a popular conventional upsample method in image super-resolution \cite{DBLP:conf/eccv/DongLT16,DBLP:journals/corr/MaoSY16a,DBLP:conf/iccv/0001LLG17}.
However, they equally treat high-frequency and low-frequency information. Therefore high-frequency details are hard to be fully explored to upscale.
Moreover, according to our experiments, we find the deconvolution easily destabilizes training process.
To solve these issues, we separately upscale high-frequency and low-frequency information by a pyramid structure for upsampling.
\begin{figure}[h]
\centering
\includegraphics[width=0.95\linewidth]{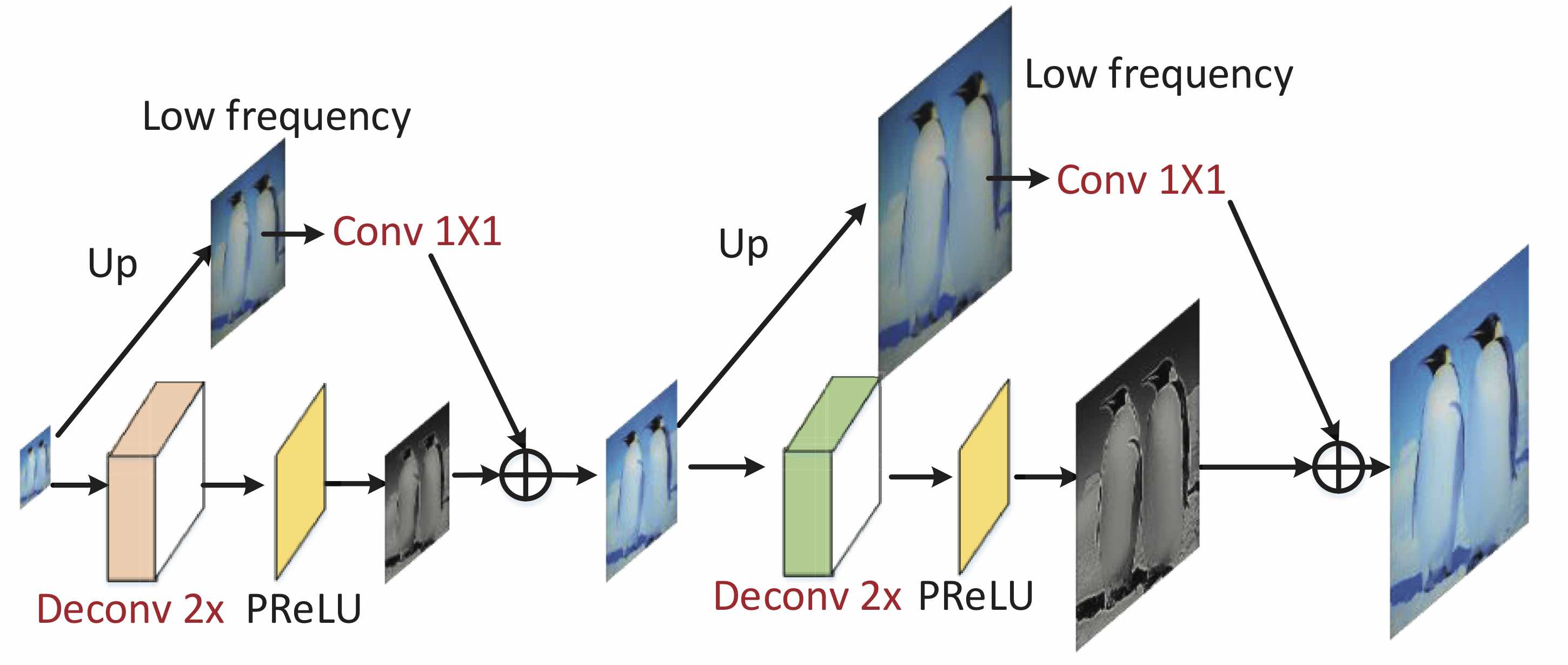}
\caption{\label{RDeConv}%
           Structure of the residual deconvolution strategy, the red parts are trainable. $\oplus$ denotes element-wise addition.}
\end{figure} 

As shown in Figure \ref{RDeConv}, the structure consists two blocks. In each block, it contains a deconvolution layer, a PReLU, and a $1 \times 1$ convolution layer. We use a ``nearest" interpolation function $Up(\cdot)$ and the $1 \times 1$ convolution $W_{1\times1}$ to upsample low-frequency information, it can be formulated as:
\begin{equation}
x_{low}  = W_{1\times1}*Up(x_{in}) 
\end{equation}
where ``$*$" denotes convolutional operation, and $x_{in}$ is the input feature map.
In addition, the deconvolution layer and the PReLU can upsample high-frequency of the feature map by $2 \times$ in each block:
\begin{equation}
x_{high} = \mathrm{PReLU}(W_{deconv}*x_{in})
\end{equation}
where $W_{deconv}$ denotes the deconvolution layer's operation. We perform element-wise addition for $x_{high}$ and $x_{low}$, thus we get final upsampled output of each building block.
Notably, the input and output channel should be equal, and the two deconvolution layers have different weights. 

\section{Experiment}
\label{experiment}
\subsection{Data and evaluation metrics}
We follow work \cite{DBPN} to train our network using high-quality (2K resolution) DIV2K dataset \cite{DBLP:conf/cvpr/TimofteAG0ZLSKN17} and ImageNet dataset \cite{DBLP:conf/cvpr/DengDSLL009}. Data augmentation is adopted with random flip, rotation ($90^\circ$, $180^\circ$, and $270^\circ$).
To evaluate our method, four benchmark datasets are adopted: Set5 \cite{DBLP:conf/bmvc/BevilacquaRGA12}, Set14 \cite{DBLP:conf/cas/ZeydeEP10}, BSD100 \cite{DBLP:conf/iccv/MartinFTM01} and Urban100 \cite{DBLP:conf/cvpr/HuangSA15} datasets. Set5 \cite{DBLP:conf/bmvc/BevilacquaRGA12} and Set14 \cite{DBLP:conf/cas/ZeydeEP10} contain $5$ and $14$ different types of images, respectively. BSD100 includes $100$ natural images. And the Urban100 contains $100$ images of urban scenario. All experiments are performed using a $4 \times$ up-scaling factor from low resolution to high resolution. The peak signal-to-noise ratio (PSNR) and structural similarity (SSIM) index are two criterion metrics for evaluation. The PSNR and SSIM are calculated on the Y-channel of images.

\subsection{Ablation investigation}
We build a lightweight ADRD architecture to evaluate each proposed module. It contains 4 dense block with $6$, $10$, $14$, and $10$ dense layers. Experiments adopt $40 \times 40$ patches for training, and other settings are same as Section \ref{Implementation}.
\paragraph{WDB evaluation.} We investigate WDB and with different growth rates $G$ ($12$, $24$, $48$). To verify the effectiveness of WDB, the experiment compares it with dense block (DB) which weights are fixed and equal to $1$. As shown in Table \ref{tab:ABCcompWDB}, by adopting a group of trainable weights, WDB can consistently achieve higher scores than DB when setting different growth rates. It achieves more apparent PSNR promotion with growth rates increased.
\begin{table} [h]
\small
\begin{center}
\setlength{\tabcolsep}{0.9 mm}
\begin{tabular}{|l|cccccc|}
\hline
 Index & DB-$12$ & WDB-$12$ & DB-$24$ & WDB-$24$ & DB-$48$ & WDB-$48$ \\
\hline\hline
PSNR & $31.27$ & $31.28$ &  $31.65$ & $31.71$ & $31.82$ & $31.89$ \\
SSIM  & $0.8911$ & $0.8916$ &  $0.8956$ & $0.8957$ & $0.8970$ &  $0.8975$ \\
\hline
\end{tabular}
\end{center}
\setlength{\abovecaptionskip}{0pt}
\caption{\label{tab:ABCcompWDB}%
		Investigations of WDB with different growth rates on Set5 with scaling factor $4$.}
\end{table}
\vspace{-0.5 cm}
\begin{figure}[h]
\centering
\includegraphics[width=1.0\linewidth]{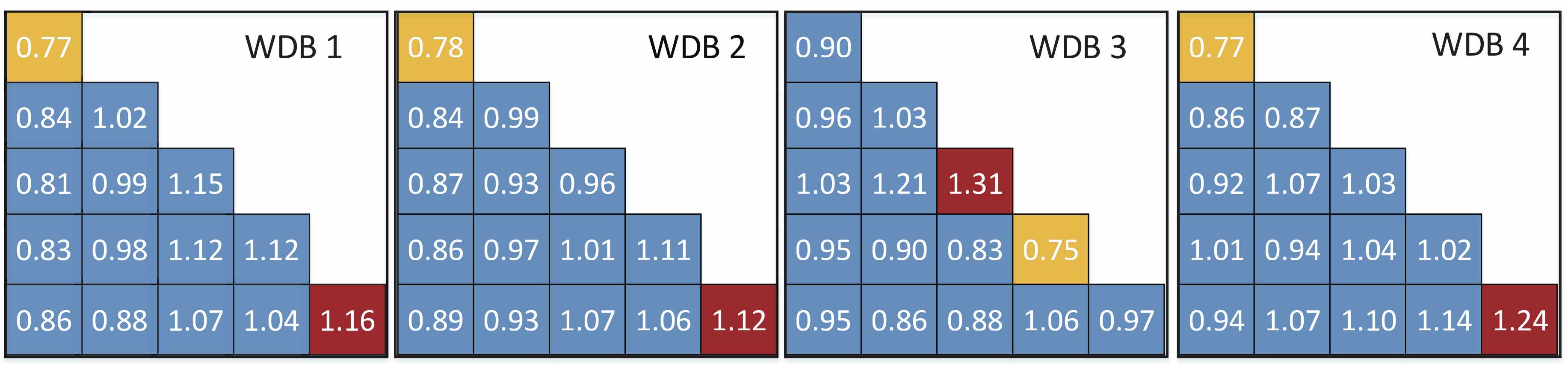}
\caption{\label{weighted}%
           Weight matrix of different blocks, the foregoing five dense layers are selected for exhibition.}
\end{figure} 

An example of weighted matrixes of WDB is shown in Figure \ref{weighted} which shows weights of the foremost five layers. The red part in each dense block is the maximum weight and the yellow one is the minimum weight. The minimum value for the $1$-th, $2$-th, $4$-th dense block exists in the head layer while the biggest value comes from the nearest layer. As for the $3$-th block, the maximum and minimum values both appear in the nearest layer. It reveals that the weights of the nearest features are more sensitive and important than the preceding levels. Conclusively, WDB can learn meaningful weights adaptively from training data.

\begin{figure}[h]
\centering
\includegraphics[width=1.0\linewidth]{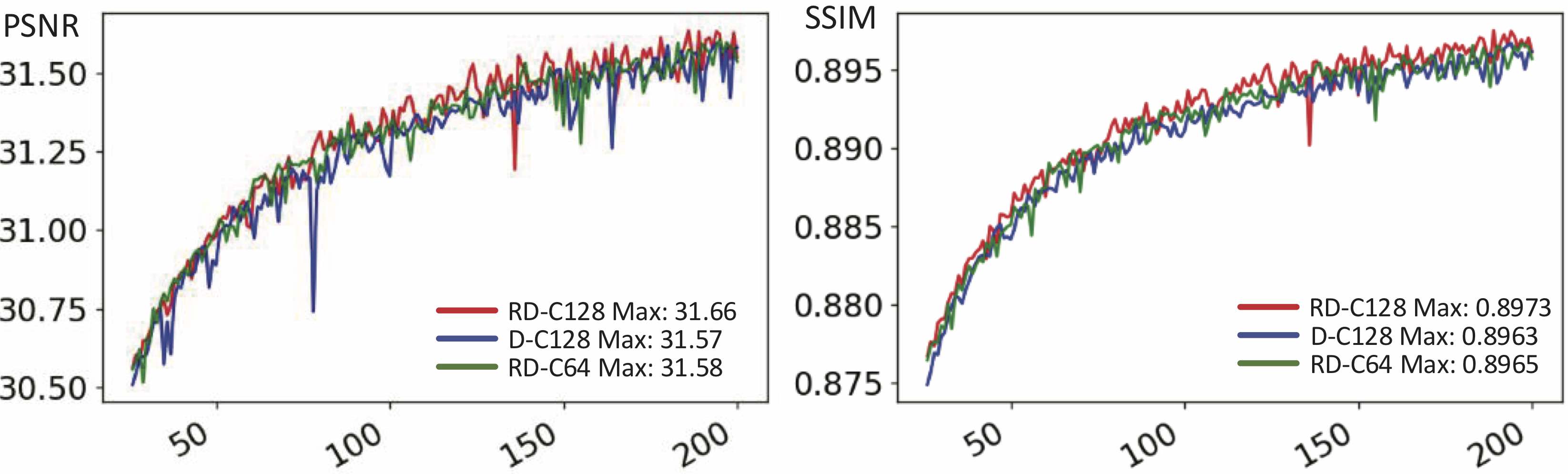}
\caption{\label{traindeconv} Curve convergence of PSNR and SSIM on Set5.}
\end{figure}

%\begin{table} [h]
%\small
%\begin{center}
%\setlength{\tabcolsep}{0.9 mm}
%\begin{tabular}{|l|c|ccccccc|}
%\hline
% Index &N& A & B & C & AB & AC & BC & ABC\\
%\hline\hline
%PSNR & $31.70$ & $31.77$ & $31.79$ &  $31.75$ & $31.84$ & $31.83$ & $31.81$ & $\textbf{31.87}$ \\
%SSIM &  $0.894$ & $0.895$ & $0.895$ &  $0.894$ & $0.897$ & $0.896$ &  $0.896$ & $\textbf{0.897}$ \\
%\hline
%\end{tabular}
%\end{center}
%\caption{\label{tab:ABCcomp1}%
%		Investigations with different proposed methods on Set5 with scaling factor 4.}
%\end{table}
\paragraph{Spatial attention evaluation.} We adopt growth rate $G$ = $16$, $G$ = $20$ and $32$ to verify the effectiveness of SA. ``noSA" denotes there is no SA in the network. Except for PSNR and SSIM evaluation, we introduce relative content increasing rate (RCIR) to verify the ability of SA module for enhancing high-frequency features. According to \cite{DBLP:conf/cvpr/LedigTHCCAATTWS17}, they utilized a pre-trained VGG network to optimize the content loss to make super-resolved images have more high-frequency information. We use this property to calculate RCIR. Firstly, we calculate the mean absolute error (MAE) between HR and interpolated images based on the content:
\begin{equation}
\mathrm{E_{HR-Bic} = \mathrm{MAE}(\phi_{VGG}(I^{HR}) - \phi_{VGG}(I^{Bic}))},
\end{equation}
where $\mathrm{\phi}$ is the $31$-th layer's output from the VGG16, $\mathrm{I^{HR}}$ is high-resolution image. Then, the MAE between HR and super-resolved images is calculated:
\begin{equation}
\mathrm{E_{HR-SR} = MAE(\phi_{VGG}(I^{HR}) - \phi_{VGG}(I^{SR}))}.
\end{equation}
where $\mathrm{I^{SR}}$ is a super-resolved image. We assume that $\mathrm{E_{HR-Bic}}$ is larger than $\mathrm{E_{HR-SR}}$, and the value of $\mathrm{E_{HR-Bic}}$ is not equal to zero. At last, the $\mathrm{S}$ can be calculated as:
\begin{equation}
 \mathrm{S = 1- E_{HR-SR}/E_{HR-Bic}}.
\end{equation}
A model can achieve high RCIR when it has relatively low error between HR and SR.
\begin{table} [h]
\small
\begin{center}
\setlength{\tabcolsep}{0.8 mm}
\begin{tabular}{|l|cccccc|}
\hline
 Index &noSA-16 & SA-16 & noSA-20 &SA-20 & noSA-32 & SA-32\\
\hline\hline
PSNR & $28.25$ & $28.39$&$28.28$ & $28.43$ & $28.43$ &  $28.55$  \\
SSIM &  $0.7784$ &$0.7820$& $0.7788$&  $0.7830$ & $0.7827$ &  $0.7848$  \\
RCIR &  $0.172$ & $0.175$ & $0.173$  & $0.183$ &$0.181$ &  $0.190$  \\
\hline
\end{tabular}
\end{center}
\setlength{\abovecaptionskip}{0pt}
\caption{\label{tab:SACMP}%
		Evaluation of SA with different growth rates on Set14 with $4 \times$ up-scaling factor.}
\end{table}
As shown in Table \ref{tab:SACMP}, SA improves PSNR more than $0.1db$ in each growth rate, and it increases output RCIR with a large extent, thus high-frequency details of an image tend to be recovered more clearly.
Besides, compared SA ($G$ = $20$) with noSA ($G$ = $32$), they have the almost same parameters ($1.52$M and $1.54$M), but SA still achieves higher RCIR and SSIM than no SA.
Notably, increasing $G$ can achieve better performance. However, it will construct a wide network and brings severalfold computational load, so utilizing SA modules is an effective way to boost image SR performance without too much computational load.
\paragraph{Residual deconvolution evaluation.} Residual deconvolution (RD) strategy bypasses low-frequency and focus on high-frequency deconvolution. Here, we take WDB with16-growth rate and utilize SA module to exhibit training curves of deconvolution (denotes D) and RD, as shown in Figure \ref{traindeconv}.

 \newcommand{\tabincell}[2]{
\begin{tabular}{@{}#1@{}}#2\end{tabular}
} 
\begin{table*} [t]
\small  
\begin{center}
\setlength{\tabcolsep}{1.6 mm}
\begin{tabular}{|l|c|cccccccccc|}
\hline
 \tabincell{c}{\ \ Dataset } & Index & \tabincell{c}{Bicubic} & \tabincell{c}{ A+ }  & \tabincell{c}{SRCNN} & \tabincell{c}{VDSR} & \tabincell{c}{LapSRN} & \tabincell{c}{SRDense}& \tabincell{c}{SR-DDNet} &  \tabincell{c}{\ RDN \ }  & \tabincell{c}{D-DBPN} & \tabincell{c}{ \ ADRD \  } \\
\hline\hline
\ \ \ \ Set5 & PSNR &$28.42$ &$30.28$ & $30.48$  & $31.35$ & $31.54$  & $32.02$ & $32.21$  &$32.47$ & $\textbf{32.47}$   & $32.45$  \\
& SSIM  &$0.8104$ &$0.8603$ & $0.8820$  & $0.8855$ & $0.8934$  & $0.8982$ & $0.8988$  & $0.8990$ & $0.8980$ & $\textbf{0.8999}$   \\
\hline
\ \ \ Set14 & PSNR  &$26.00$ &$27.32$ & $27.50$  & $28.03$ & $28.19$  & $28.50$ & $28.71$   &$28.81$  & $28.82$ & $\textbf{28.84}$ \\
& SSIM &$0.7027$ &$0.7491$ & $0.7513$  & $0.7701$ & $0.7720$  & $0.7782$  & $0.7805$ &   $0.7871$  &  $0.7861$ & $\textbf{0.7923}$   \\
\hline
\  BSD100 & PSNR&$25.96$ &$26.82$ & $26.90$  & $27.29$ & $27.32$  & $27.53$ & $27.69$  & $27.72$ & $\textbf{27.72}$  & $27.69$   \\
&SSIM  &$0.6675$ &$0.7087$ & $0.7101$  & $0.7264$ & $0.7280$  & $0.7337$ & $0.7396$ & $0.7419$  & $0.7401$& $\textbf{0.7477}$   \\
\hline
Urban100 & PSNR&$23.14$ &$24.32$ & 24.52  & $25.18$ & $25.21$  & $26.05$ & $26.21$  & $26.61$  & \ \ $27.08^{\star}$ & $\textbf{\ \ 27.26}^{{\star}}$   \\
& SSIM&$0.6577$ &$0.7183$ &$0.7221$  & $0.7553$ & $0.7561$  & $0.7819$  & $0.7884$ & $0.8028$  & $0.7972$ & $\textbf{0.8041}$   \\
\hline
\end{tabular}
\end{center}
\setlength{\abovecaptionskip}{0pt}
\caption{\label{fulltable}%
Comparisons with the state-of-the-art methods by PSNR and SSIM ($4\times$). Scores in bold denote the highest values ($^{\star}$ indicates that the input is divided into four parts and calculated due to computation limitation of large size images).}
\end{table*}
The C$128$ denotes $128$-channel features. Compared with deconvolution, RD can not only make the network achieve better results, but also stabilize the training process. Because it can reduce the influence of low frequency. Though RD-C$64$ has only 64 channels, but it acquires comparable performance with D-C$128$, showing superiority of the proposed strategy.

\subsection{Comparisons with the state-of-the-arts}
We compare ADRD with state-of-the-art methods, as shown in Table \ref{fulltable}. Here, the bicubic interpolation is viewed as a baseline for comparisons. A+ \cite{DBLP:conf/iccv/TimofteDG13} is introduced as a conventional machine learning approach. Some CNN based methods, i.e. SRCNN \cite{DBLP:journals/pami/DongLHT16}, VDSR \cite{DBLP:conf/cvpr/KimLL16a}, LapSRN \cite{DBLP:conf/cvpr/LaiHA017}, and D-DBPN \cite{DBPN} are introduced. 
SRDenseNet \cite{DBLP:conf/iccv/0001LLG17} (denotes as SRDense), SR-DDNet \cite{SRDDNet}, and RDN \cite{RDN} are three different sizes of dense block based networks are also cited in the comparison list.
ADRD achieves the highest SSIM among all methods, it tends to have better quality in human perception \cite{SSIM}. Because ADRD is adept at recovering high-frequency information.
Additionally, ADRD also outperforms D-DBPN nearly $0.2db$ PSNR on Urban100 dataset that contains many large-size real-world images.

 \begin{figure}[h]
  \centering
  \includegraphics[width=0.9\linewidth]{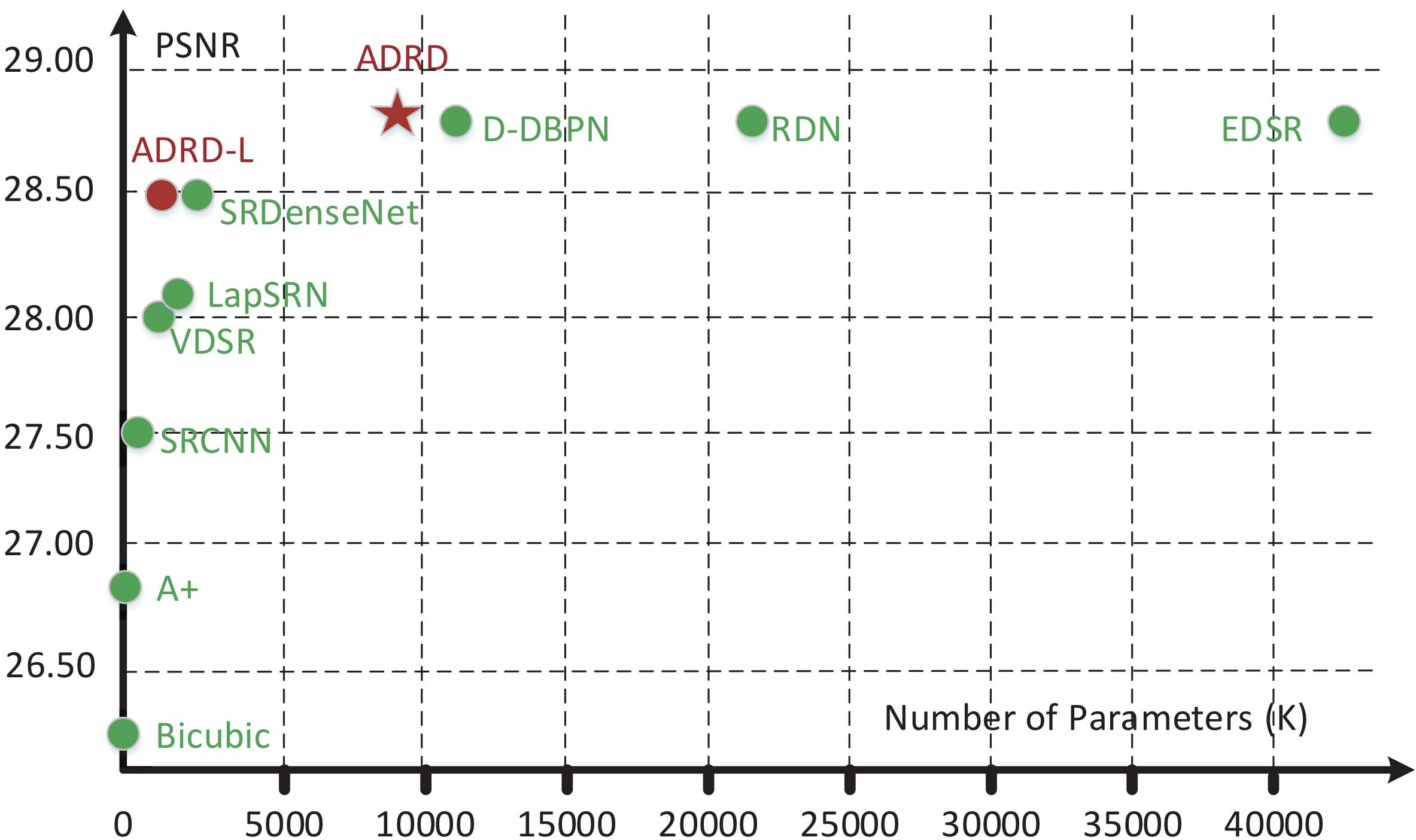}
  \caption{\label{param}%
           Parameters and PSNR comparison on Set14.}
\end{figure} 

\begin{figure*}[t]
\centering
%-------------------------------------------------------------------------
%-------------------------------------------------------------------------
\begin{minipage}[t]{0.12\linewidth}
\centering
\includegraphics[width=1\linewidth]{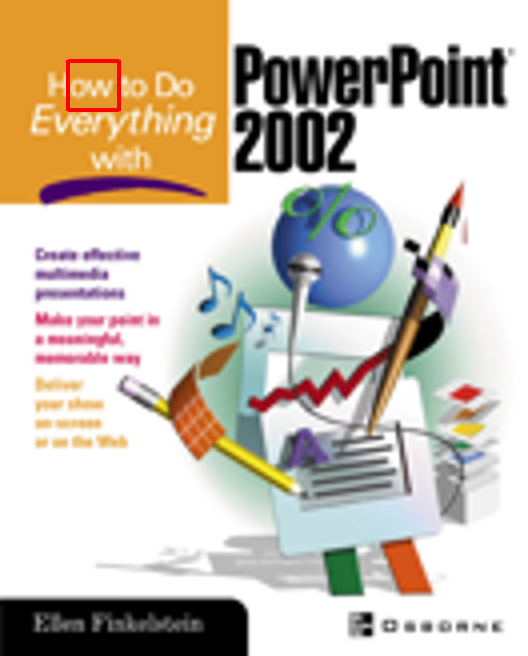}
\end{minipage}
\begin{minipage}[t]{0.12\linewidth}
\centering
\includegraphics[width=1\linewidth]{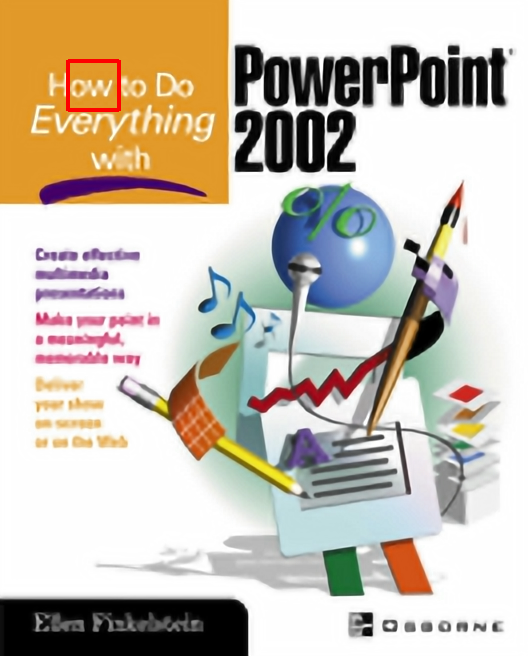}
\end{minipage}
\begin{minipage}[t]{0.12\linewidth}
\centering
\includegraphics[width=1\linewidth]{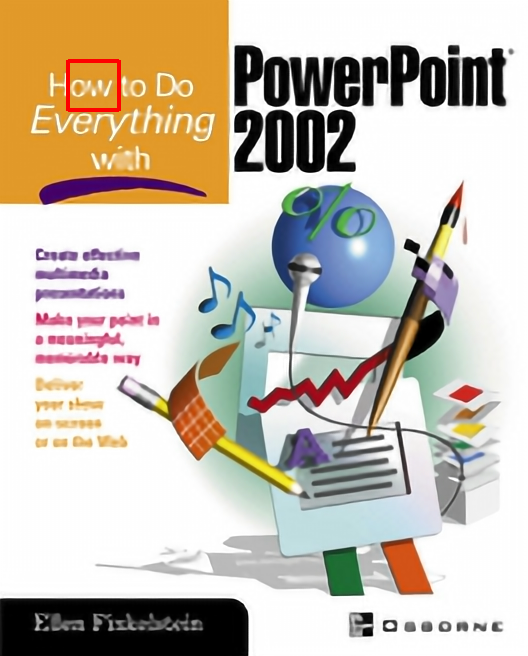}
\end{minipage}
\begin{minipage}[t]{0.12\linewidth}
\centering
\includegraphics[width=1\linewidth]{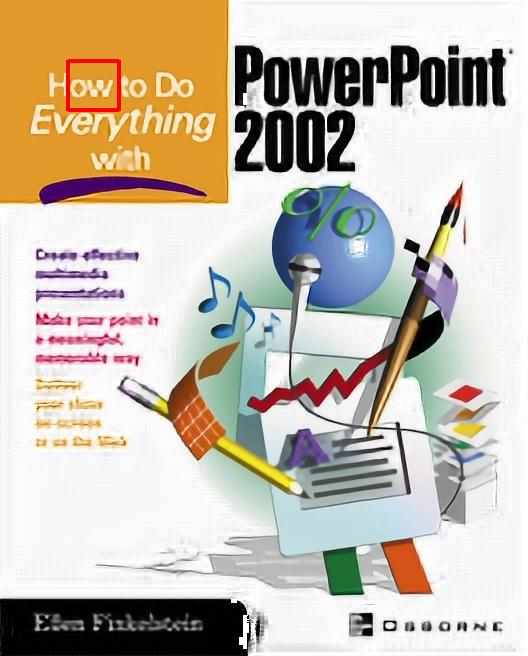}
\end{minipage}
\begin{minipage}[t]{0.12\linewidth}
\centering
\includegraphics[width=1\linewidth]{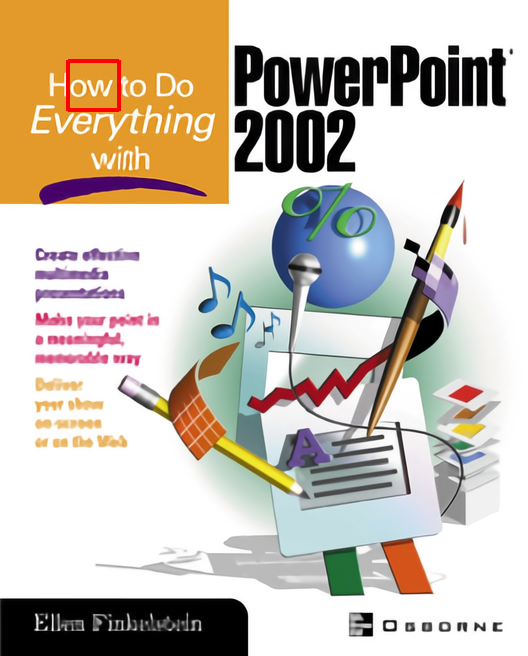}
\end{minipage}
\begin{minipage}[t]{0.12\linewidth}
\centering
\includegraphics[width=1\linewidth]{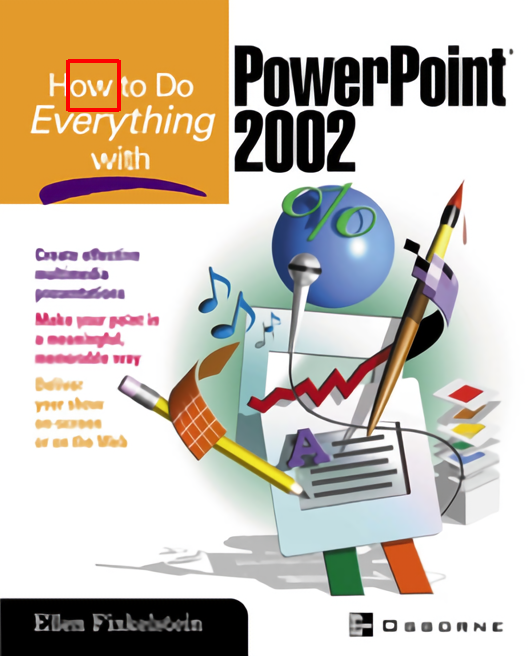}
\end{minipage}
\begin{minipage}[t]{0.12\linewidth}
\centering
\includegraphics[width=1\linewidth]{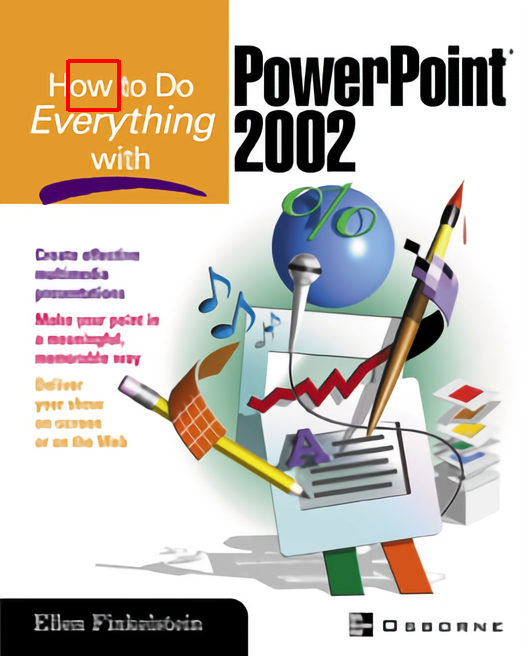}
\end{minipage}
\begin{minipage}[t]{0.12\linewidth}
\centering
\includegraphics[width=1\linewidth]{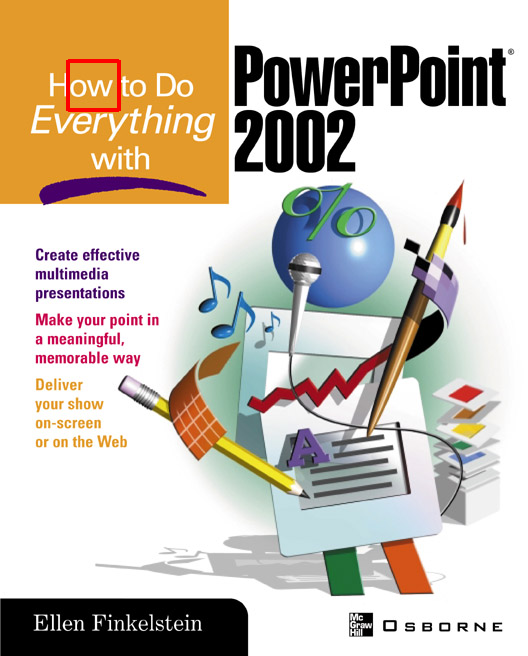}
\end{minipage}
%-------------------------------------------------------------------------
\begin{minipage}[t]{0.12\linewidth}
\centering
\includegraphics[width=1\linewidth]{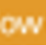}

\end{minipage}
\begin{minipage}[t]{0.12\linewidth}
\centering
\includegraphics[width=1\linewidth]{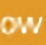}

\end{minipage}
\begin{minipage}[t]{0.12\linewidth}
\centering
\includegraphics[width=1\linewidth]{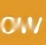}

\end{minipage}
\begin{minipage}[t]{0.12\linewidth}
\centering
\includegraphics[width=1\linewidth]{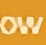}

\end{minipage}
\begin{minipage}[t]{0.12\linewidth}
\centering
\includegraphics[width=1\linewidth]{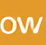}

\end{minipage}
\begin{minipage}[t]{0.12\linewidth}
\centering
\includegraphics[width=1\linewidth]{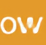}

\end{minipage}
\begin{minipage}[t]{0.12\linewidth}
\centering
\includegraphics[width=1\linewidth]{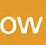}

\end{minipage}
\begin{minipage}[t]{0.12\linewidth}
\centering
\includegraphics[width=1\linewidth]{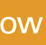}

\end{minipage}

\begin{minipage}[t]{0.12\linewidth}
\centering
\includegraphics[width=1\linewidth]{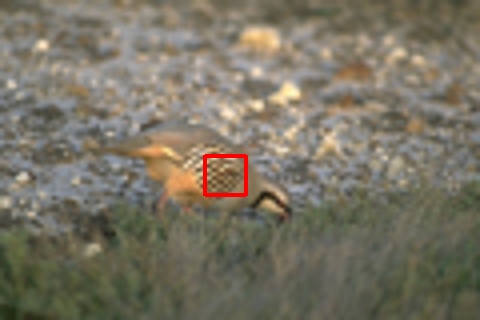}
\end{minipage}
\begin{minipage}[t]{0.12\linewidth}
\centering
\includegraphics[width=1\linewidth]{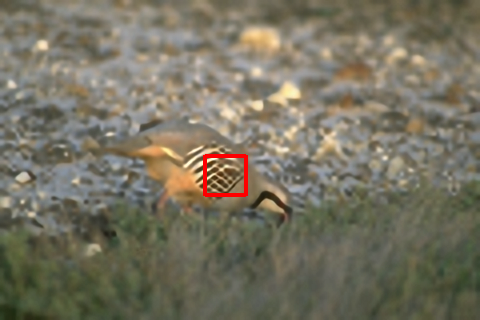}
\end{minipage}
\begin{minipage}[t]{0.12\linewidth}
\centering
\includegraphics[width=1\linewidth]{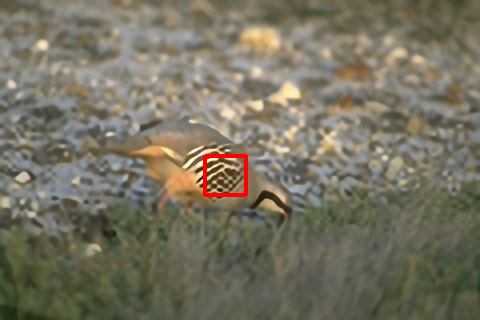}
\end{minipage}
\begin{minipage}[t]{0.12\linewidth}
\centering
\includegraphics[width=1\linewidth]{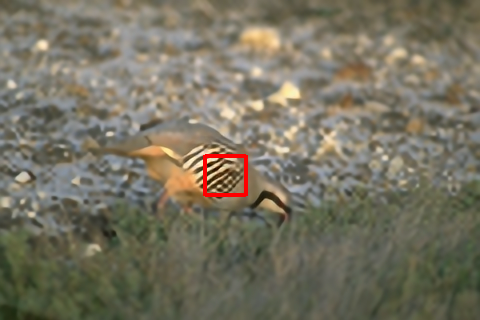}
\end{minipage}
\begin{minipage}[t]{0.12\linewidth}
\centering
\includegraphics[width=1\linewidth]{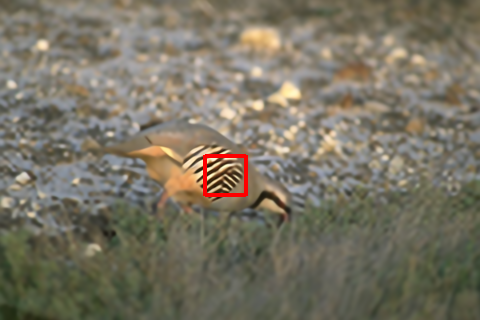}
\end{minipage}
\begin{minipage}[t]{0.12\linewidth}
\centering
\includegraphics[width=1\linewidth]{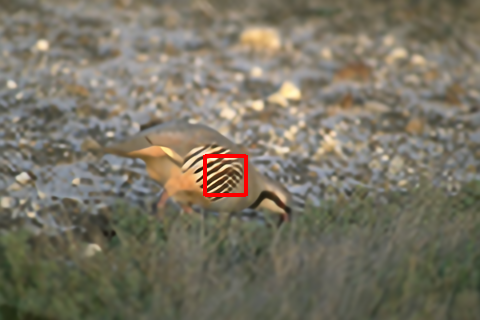}
\end{minipage}
\begin{minipage}[t]{0.12\linewidth}
\centering
\includegraphics[width=1\linewidth]{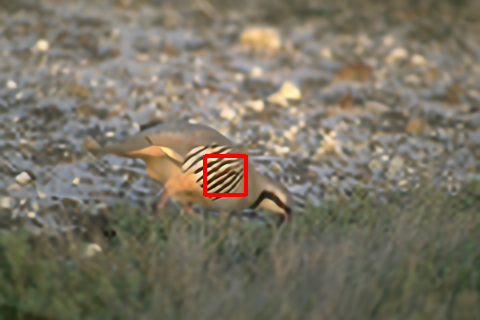}
\end{minipage}
\begin{minipage}[t]{0.12\linewidth}
\centering
\includegraphics[width=1\linewidth]{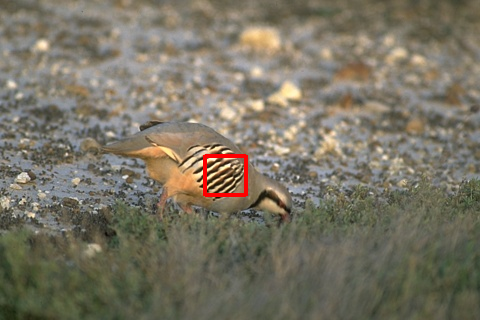}
\end{minipage}
%-------------------------------------------------------------------------
\begin{minipage}[t]{0.12\linewidth}
\centering
\includegraphics[width=1\linewidth]{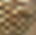}
\small{Bicubic}
\end{minipage}
\begin{minipage}[t]{0.12\linewidth}
\centering
\includegraphics[width=1\linewidth]{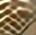}
\small{VDSR}
\end{minipage}
\begin{minipage}[t]{0.12\linewidth}
\centering
\includegraphics[width=1\linewidth]{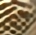}
\small{LapSRN}
\end{minipage}
\begin{minipage}[t]{0.12\linewidth}
\centering
\includegraphics[width=1\linewidth]{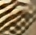}
\small{SRDenseNet}
\end{minipage}
\begin{minipage}[t]{0.12\linewidth}
\centering
\includegraphics[width=1\linewidth]{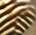}
\small{RDN}
\end{minipage}
\begin{minipage}[t]{0.12\linewidth}
\centering
\includegraphics[width=1\linewidth]{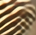}
\small{D-DBPN}
\end{minipage}
\begin{minipage}[t]{0.12\linewidth}
\centering
\includegraphics[width=1\linewidth]{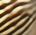}
\small{\textbf{ADRD}}
\end{minipage}
\begin{minipage}[t]{0.12\linewidth}
\centering
\includegraphics[width=1\linewidth]{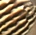}
\small{HR}
\end{minipage} \\
\caption{\label{imageCompare}%
            Visual comparisons with up-scaling factor $4\times$. From top to bottom: ``ppt3" from Set14 and ``img\_093" from BSD100.}
\end{figure*}

Comprehensively, we visualize parameters and PSNR comparisons on the Set14 dataset ($4 \times$). As shown in Figure \ref{param}, ADRD has less (about 9700 K) a half parameters than RDN, but it still shows a bit promotion. ADRD also outperforms dense block based network, i.e. SRDenseNet, SR-DDNet more than $0.3db$ and $0.1db$, demonstrating the superiority of our method.
Visual comparisons are shown in Figure \ref{imageCompare}, in the first group comparison, ADRD clearly recovers the word ``W"  but others exist breakage. The second group shows strong recovery capability of ADRD in textures, which is close to the HR image.

\subsection{Robustness comparison}
The robustness is also essential for image super-resolution. We evaluate our method on different Gaussian noise levels. Here, four kinds of noise variances are used: $5 \times 10^{-5}$,  $1 \times 10^{-4}$, $2 \times 10^{-4}$, and $5 \times 10^{-4}$. The Bicubic is viewed as the baseline. Three state-of-the-art networks D-DBPN \cite{DBPN}, RDN \cite{RDN}, LapSRN \cite{DBLP:conf/cvpr/LaiHA017} are introduced for comparisons. The detailed results are shown in Table \ref{tab:noisecompare}.

\begin{table} [h]
\small
\begin{center}
\setlength{\tabcolsep}{0.9 mm}
\begin{tabular}{|l|c|c|c|c|c|}
\hline
\ \ Level  &Bicubic & LapSRN &\   RDN \   & D-DBPN  & ADRD \\
\hline\hline
$5 \times 10^{-5}$  & $28.38$ & $30.84$ &$31.82$  & $31.86$  & $\textbf{31.90}$  \\
\hline
$1 \times 10^{-4} $  & $28.35$& $30.66$ &$31.44$   & $31.45$ & $\textbf{31.47}$  \\
\hline
$2 \times 10^{-4}$  & $28.27$& $30.24$ & $30.77$  & $30.86$ & $\textbf{30.86}$  \\
\hline
$5 \times 10^{-4}$  & $28.04$& $29.31$ & $29.55$ &  $29.55$ & $\textbf{29.69}$  \\
\hline
\end{tabular}
\end{center}
\setlength{\abovecaptionskip}{0pt}
\caption{\label{tab:noisecompare}%
 PSNR results of different noise levels on Set5.}
\end{table}

\begin{figure}[h]
\centering
\setlength{\abovecaptionskip}{4pt}
%-------------------------------------------------------------------------
\begin{minipage}[t]{0.155\linewidth}
\centering
\includegraphics[width=1\linewidth]{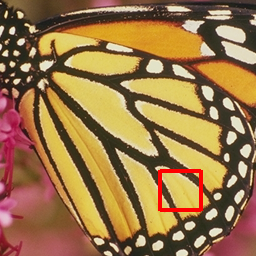}

\end{minipage}
\begin{minipage}[t]{0.155\linewidth}
\centering
\includegraphics[width=1\linewidth]{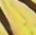}

\end{minipage}
\begin{minipage}[t]{0.155\linewidth}
\centering
\includegraphics[width=1\linewidth]{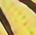}

\end{minipage}
\begin{minipage}[t]{0.155\linewidth}
\centering
\includegraphics[width=1\linewidth]{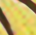}

\end{minipage}
\begin{minipage}[t]{0.155\linewidth}
\centering
\includegraphics[width=1\linewidth]{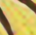}

\end{minipage}
\begin{minipage}[t]{0.155\linewidth}
\centering
\includegraphics[width=1\linewidth]{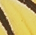}
\end{minipage}
%-------------------------------------------------------------------------
\begin{minipage}[t]{0.155\linewidth}
\centering
\includegraphics[width=1\linewidth]{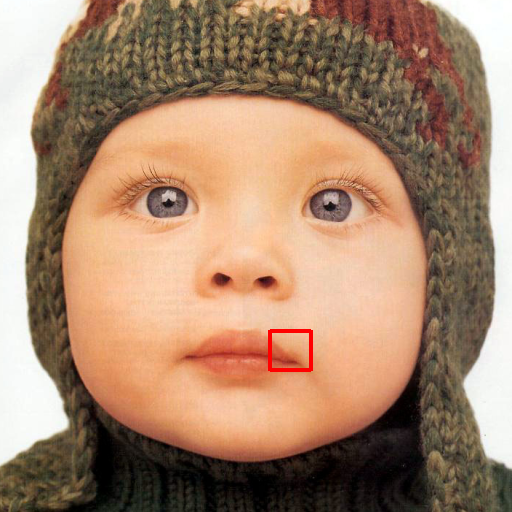}
\end{minipage}
\begin{minipage}[t]{0.155\linewidth}
\centering
\includegraphics[width=1\linewidth]{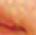}
\small{LapSRN}
\end{minipage}
\begin{minipage}[t]{0.155\linewidth}
\centering
\includegraphics[width=1\linewidth]{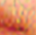}
\small{RDN}
\end{minipage}
\begin{minipage}[t]{0.155\linewidth}
\centering
\includegraphics[width=1\linewidth]{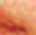}
\small{D-DBPN}
\end{minipage}
\begin{minipage}[t]{0.155\linewidth}
\centering
\includegraphics[width=1\linewidth]{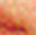}
\small{\textbf{ADRD}}
\end{minipage}
\begin{minipage}[t]{0.155\linewidth}
\centering
\includegraphics[width=1\linewidth]{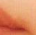}
\small{HR}
\end{minipage}\\

\caption{\label{fignoise}%
		Visual comparison of Set5 on $5 \times 10^{-4}$ noise.}
\end{figure}

\vspace{-0.2 cm}
ADRD outperforms all other methods in each noise level. Though RDN is also a dense block based network, it is easy to be attacked by noises. Despite D-DBPN surpasses ADRD on Set5 in PSNR as shown in Table \ref{fulltable}, it is lower than ours in the noise conditions.
Visual comparisons under the $5 \times 10^{-4}$ noise level are shown in Figure \ref{fignoise}. ADRD has less damage in local details. It is mainly due to the attention mechanism can reduce the weights for some noisy features by attentive maps. Therefore, ADRD is not only an effective model, but also a robust one, showing superior anti-noise capability. 

\vspace{-0.2 cm}
\subsection{Implementation details}
\label{Implementation}
\paragraph{Network setting.}  The final ADRD is trained specially for a $4 \times$ scale factor super-resolution. The primary convolution is composed of a $3 \times 3$ convolutional layer and a ReLU. The proposed ADRD model contains $4$ WDBs with $6$, $12$ , $48$ and $32$ dense layers, respectively. It utilizes 32-channel primary features, and the growth rate of WDB is set to $32$. The $\lambda$ of SA is set to $0.5$ and the channel number of global bottleneck layer is $256$.
In our network, the sizes of the convolutional filters are set to $3 \times 3$ and $1 \times 1$. For $3 \times 3$ convolutional filters, the padding is set to $1$. Notably, there is no batch normalization in ADRD, because it removes the range flexibility of the features \cite{DBPN}.
\paragraph{Training detail.} We randomly crop a set of $200 \times 200$  patches for training, thus the size of LR patch is $50 \times 50$. The training batch size is set to $16$ in each back-propagation. All the weights of weighted dense connections are initialized by $1$. 
This network is trained via pixel-wise mean square error (MSE) loss between super-resolved HR images and ground-truth HR images. The Adam \cite{DBLP:journals/corr/KingmaB14} is adopted for optimizing ADRD, and the initial learning rate is set to $0.0001$. For each $200$ epochs, the learning rate will decrease by the scale of $0.5$. After $500$ epochs, we randomly select 50000 images from ImageNet to fine-tune our networks using $30 \times 30$ patch size.
Experiments are performed on two NVIDIA Titan Xp GPUs for training and testing. The training process costs about $48$ hours for $200$ epochs, and the average testing speed of an image on Set5 dataset is $0.17$ s.

\subsection{Application for recognition}
ADRD is also beneficial for low-resolution image recognition. Here, we conduct the experiment on a real-world Pairs \& Oxford dataset \cite{DBLP:conf/cvpr/PhilbinCISZ07,DBLP:conf/cvpr/PhilbinCISZ08} that totally contains  $29$ categories. A VGG16 is trained on the dataset. Then, we adopt different models to super resolve LR testing images. The super-resolved testing images will be fed into the VGG network to test recognition accuracy. 

As shown in Table \ref{tab:class}, ADRD promotes $2.3 \%$ Top-1 accuracy, while the RDN only promotes $1.2 \%$. The results demonstrate that ADRD is good at real-world image super-resolution. As shown in Figure \ref{figclass}, the super-resolved images have clear textures, and conform to human perception.
\vspace{-0.15 cm}
\begin{table} [h]
\small
\begin{center}
\setlength{\tabcolsep}{1.0 mm}
\begin{tabular}{|c|ccccc|}
\hline
Acc (\%) & Bicubic & LapSRN & RDN & D-DBPN & ADRD \\
\hline\hline
Top-1 & $53.4$ & $52.1$  &  $54.6$  & $55.1$& $\textbf{55.7}$ \\
Top-5 & $82.5$ & $82.5$  &  $83.6$  & $83.9$& $\textbf{84.2}$ \\
\hline
\end{tabular}
\end{center}
\setlength{\abovecaptionskip}{0pt}
\caption{\label{tab:class}%
	Recognition accuracy on Pairs \& Oxford.}
\end{table}

\vspace{-0.7 cm}
\begin{figure}[H]
\centering
\setlength{\abovecaptionskip}{4pt}
\begin{minipage}[t]{0.145\linewidth}
\centering
\includegraphics[width=1\linewidth]{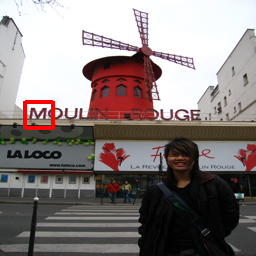}
\end{minipage}
\begin{minipage}[t]{0.155\linewidth}
\centering
\includegraphics[width=1\linewidth]{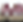}
\end{minipage}
\begin{minipage}[t]{0.155\linewidth}
\centering
\includegraphics[width=1\linewidth]{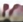}
\end{minipage}
\begin{minipage}[t]{0.155\linewidth}
\centering
\includegraphics[width=1\linewidth]{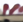}
\end{minipage}
\begin{minipage}[t]{0.155\linewidth}
\centering
\includegraphics[width=1\linewidth]{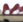}
\end{minipage}
\begin{minipage}[t]{0.155\linewidth}
\centering
\includegraphics[width=1\linewidth]{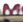}
\end{minipage}
%-------------------------------------------------------------------------
\begin{minipage}[t]{0.145\linewidth}
\centering
\includegraphics[width=1\linewidth]{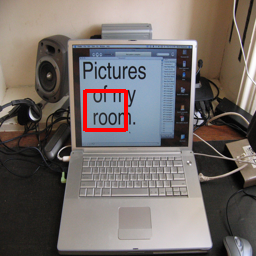}
\end{minipage}
\begin{minipage}[t]{0.155\linewidth}
\centering
\includegraphics[width=1\linewidth]{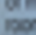}
\small{Bicubic}
\end{minipage}
\begin{minipage}[t]{0.155\linewidth}
\centering
\includegraphics[width=1\linewidth]{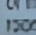}
\small{LapSRN}
\end{minipage}
\begin{minipage}[t]{0.155\linewidth}
\centering
\includegraphics[width=1\linewidth]{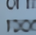}
\small{RDN}
\end{minipage}
\begin{minipage}[t]{0.155\linewidth}
\centering
\includegraphics[width=1\linewidth]{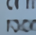}
\small{D-DBPN}
\end{minipage}
\begin{minipage}[t]{0.155\linewidth}
\centering
\includegraphics[width=1\linewidth]{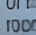}
\small{\textbf{ADRD}}
\end{minipage}
\caption{\label{figclass}%
		Visual results of different super-resolution approaches.}
\end{figure}

\section{Conclusion}
\label{conclusion}
We propose a novel attention based DenseNet with a residual deconvolution for image super-resolution. In our framework, a weighted dense block is proposed to weight all the features from all preceding layers into current layer, so as to adaptively combine informative features. A spatial attention module is presented to emphasize high-frequency information after each WDB. Besides, we exhibit a residual deconvolution strategy to focus on high-frequency upsampling. Experimental results conducted on benchmark datasets demonstrate that ADRD achieves state-of-the-art performance. Our future works will concentrate on more lightweight model design and apply to low-resolution retrieval and recognition.

\newpage

\bibliographystyle{named}
\bibliography{ijcai19}

\begin{thebibliography}{}

\bibitem[\protect\citeauthoryear{Bevilacqua and et
  al.}{2012}]{DBLP:conf/bmvc/BevilacquaRGA12}
Marco Bevilacqua and Aline~Roumy et~al.
\newblock Low-complexity single-image super-resolution based on nonnegative
  neighbor embedding.
\newblock In {\em {BMVC}}, 2012.

\bibitem[\protect\citeauthoryear{Deng \bgroup \em et al.\egroup
  }{2009}]{DBLP:conf/cvpr/DengDSLL009}
Jia Deng, Wei Dong, and Richard~Socher et~al.
\newblock Imagenet: {A} large-scale hierarchical image database.
\newblock In {\em {CVPR}}, 2009.

\bibitem[\protect\citeauthoryear{Dong \bgroup \em et al.\egroup
  }{2016a}]{DBLP:journals/pami/DongLHT16}
Chao Dong, Chen~Change Loy, Kaiming He, and Xiaoou Tang.
\newblock Image super-resolution using deep convolutional networks.
\newblock {\em {IEEE} Trans. Pattern Anal. Mach. Intell.}, 38(2):295--307,
  2016.

\bibitem[\protect\citeauthoryear{Dong \bgroup \em et al.\egroup
  }{2016b}]{DBLP:conf/eccv/DongLT16}
Chao Dong, Chen~Change Loy, and Xiaoou Tang.
\newblock Accelerating the super-resolution convolutional neural network.
\newblock In {\em {ECCV}}, 2016.

\bibitem[\protect\citeauthoryear{Freedman and
  Fattal}{2011}]{DBLP:journals/tog/FreedmanF11}
Gilad Freedman and Raanan Fattal.
\newblock Image and video upscaling from local self-examples.
\newblock {\em {ACM} Trans. Graph.}, 30(2):12:1--12:11, 2011.

\bibitem[\protect\citeauthoryear{Haris \bgroup \em et al.\egroup }{2018}]{DBPN}
Muhammad Haris, Greg Shakhnarovich, and Norimichi Ukita.
\newblock Deep back-projection networks for super-resolution.
\newblock In {\em {CVPR}}, 2018.

\bibitem[\protect\citeauthoryear{He \bgroup \em et al.\egroup }{2015}]{PReLU}
Kaiming He, Xiangyu Zhang, Shaoqing Ren, and Jian Sun.
\newblock Delving deep into rectifiers: Surpassing human-level performance on
  imagenet classification.
\newblock In {\em {ICCV}}, 2015.

\bibitem[\protect\citeauthoryear{Huang \bgroup \em et al.\egroup
  }{2015}]{DBLP:conf/cvpr/HuangSA15}
Jia{-}Bin Huang, Abhishek Singh, and Narendra Ahuja.
\newblock Single image super-resolution from transformed self-exemplars.
\newblock In {\em {CVPR}}, 2015.

\bibitem[\protect\citeauthoryear{Huang \bgroup \em et al.\egroup
  }{2017}]{DBLP:conf/cvpr/HuangLMW17}
Gao Huang, Zhuang Liu, Laurens van~der Maaten, and Kilian~Q. Weinberger.
\newblock Densely connected convolutional networks.
\newblock In {\em {CVPR}}, 2017.

\bibitem[\protect\citeauthoryear{Hui \bgroup \em et al.\egroup
  }{2018}]{Hui_2018_CVPR}
Zheng Hui, Xiumei Wang, and Xinbo Gao.
\newblock Fast and accurate single image super-resolution via information
  distillation network.
\newblock In {\em {CVPR}}, June 2018.

\bibitem[\protect\citeauthoryear{Kim and
  Kwon}{2010}]{DBLP:journals/pami/KimK10}
Kwang~In Kim and Younghee Kwon.
\newblock Single-image super-resolution using sparse regression and natural
  image prior.
\newblock {\em {IEEE} Trans. Pattern Anal. Mach. Intell.}, 32(6):1127--1133,
  2010.

\bibitem[\protect\citeauthoryear{Kim \bgroup \em et al.\egroup
  }{2016a}]{DBLP:conf/cvpr/KimLL16a}
Jiwon Kim, Jung~Kwon Lee, and Kyoung~Mu Lee.
\newblock Accurate image super-resolution using very deep convolutional
  networks.
\newblock In {\em {CVPR}}, 2016.

\bibitem[\protect\citeauthoryear{Kim \bgroup \em et al.\egroup
  }{2016b}]{DBLP:conf/cvpr/KimLL16}
Jiwon Kim, Jung~Kwon Lee, and Kyoung~Mu Lee.
\newblock Deeply-recursive convolutional network for image super-resolution.
\newblock In {\em {CVPR}}, 2016.

\bibitem[\protect\citeauthoryear{Kingma and
  Ba}{2014}]{DBLP:journals/corr/KingmaB14}
Diederik~P. Kingma and Jimmy Ba.
\newblock Adam: {A} method for stochastic optimization.
\newblock {\em CoRR}, abs/1412.6980, 2014.

\bibitem[\protect\citeauthoryear{Lai \bgroup \em et al.\egroup
  }{2017}]{DBLP:conf/cvpr/LaiHA017}
Wei{-}Sheng Lai, Jia{-}Bin Huang, and Narendra~Ahuja et~al.
\newblock Deep laplacian pyramid networks for fast and accurate
  super-resolution.
\newblock In {\em {CVPR}}, 2017.

\bibitem[\protect\citeauthoryear{Ledig \bgroup \em et al.\egroup
  }{2017}]{DBLP:conf/cvpr/LedigTHCCAATTWS17}
Christian Ledig, Lucas Theis, and Ferenc~Huszar et~al.
\newblock Photo-realistic single image super-resolution using a generative
  adversarial network.
\newblock In {\em {CVPR}}, 2017.

\bibitem[\protect\citeauthoryear{Liu \bgroup \em et al.\egroup
  }{2017}]{DBLP:conf/mm/LiuLMC17}
Wu~Liu, Xinchen Liu, Huadong Ma, and Peng Cheng.
\newblock Beyond human-level license plate super-resolution with progressive
  vehicle search and domain priori {GAN}.
\newblock In {\em ACM MM}, 2017.

\bibitem[\protect\citeauthoryear{Mao \bgroup \em et al.\egroup
  }{2016}]{DBLP:journals/corr/MaoSY16a}
Xiao{-}Jiao Mao, Chunhua Shen, and Yu{-}Bin Yang.
\newblock Image restoration using convolutional auto-encoders with symmetric
  skip connections.
\newblock {\em CoRR}, abs/1606.08921, 2016.

\bibitem[\protect\citeauthoryear{Philbin \bgroup \em et al.\egroup
  }{2007}]{DBLP:conf/cvpr/PhilbinCISZ07}
James Philbin, Ondrej Chum, and Michael~Isard et~al.
\newblock Object retrieval with large vocabularies and fast spatial matching.
\newblock In {\em {CVPR}}, 2007.

\bibitem[\protect\citeauthoryear{Philbin \bgroup \em et al.\egroup
  }{2008}]{DBLP:conf/cvpr/PhilbinCISZ08}
James Philbin, Ondrej Chum, Michael Isard, Josef Sivic, and Andrew Zisserman.
\newblock Lost in quantization: Improving particular object retrieval in large
  scale image databases.
\newblock In {\em {CVPR}}, 2008.

\bibitem[\protect\citeauthoryear{R. and et
  al.}{2001}]{DBLP:conf/iccv/MartinFTM01}
David R. and Martin et~al.
\newblock A database of human segmented natural images and its application to
  evaluating segmentation algorithms and measuring ecological statistics.
\newblock In {\em {ICCV}}, 2001.

\bibitem[\protect\citeauthoryear{Tai \bgroup \em et al.\egroup
  }{2017}]{DBLP:conf/cvpr/TaiY017}
Ying Tai, Jian Yang, and Xiaoming Liu.
\newblock Image super-resolution via deep recursive residual network.
\newblock In {\em {CVPR}}, 2017.

\bibitem[\protect\citeauthoryear{Timofte \bgroup \em et al.\egroup
  }{2013}]{DBLP:conf/iccv/TimofteDG13}
Radu Timofte, Vincent~De Smet, and Luc J.~Van Gool.
\newblock Anchored neighborhood regression for fast example-based
  super-resolution.
\newblock In {\em {ICCV}}, 2013.

\bibitem[\protect\citeauthoryear{Timofte \bgroup \em et al.\egroup
  }{2017}]{DBLP:conf/cvpr/TimofteAG0ZLSKN17}
Radu Timofte, Eirikur Agustsson, and Luc Van~Gool et~al.
\newblock {NTIRE} 2017 challenge on single image super-resolution: Methods and
  results.
\newblock In {\em {CVPR} Workshops}, 2017.

\bibitem[\protect\citeauthoryear{Tong \bgroup \em et al.\egroup
  }{2017}]{DBLP:conf/iccv/0001LLG17}
Tong Tong, Gen Li, Xiejie Liu, and Qinquan Gao.
\newblock Image super-resolution using dense skip connections.
\newblock In {\em {ICCV}}, 2017.

\bibitem[\protect\citeauthoryear{Wang \bgroup \em et al.\egroup }{2004}]{SSIM}
Zhou Wang, Alan~C. Bovik, Hamid~R. Sheikh, and Eero~P. Simoncelli.
\newblock Image quality assessment: from error visibility to structural
  similarity.
\newblock {\em {IEEE} Trans. Image Processing}, 13(4):600--612, 2004.

\bibitem[\protect\citeauthoryear{Yang \bgroup \em et al.\egroup
  }{2013}]{DBLP:conf/cvpr/YangLC13}
Jianchao Yang, Zhe Lin, and Scott Cohen.
\newblock Fast image super-resolution based on in-place example regression.
\newblock In {\em {CVPR}}, 2013.

\bibitem[\protect\citeauthoryear{Zeyde and et
  al.}{2010}]{DBLP:conf/cas/ZeydeEP10}
Roman Zeyde and Michael~Elad et~al.
\newblock On single image scale-up using sparse-representations.
\newblock In {\em International Conference on Curves and Surfaces}, 2010.

\bibitem[\protect\citeauthoryear{Zhang \bgroup \em et al.\egroup }{2018}]{RDN}
Yulun Zhang, Yapeng Tian, and Yu~Kong et~al.
\newblock Residual dense network for image super-resolution.
\newblock In {\em {CVPR}}, 2018.

\bibitem[\protect\citeauthoryear{Zhu \bgroup \em et al.\egroup
  }{2018}]{SRDDNet}
Xiaobin Zhu, Zhuangzi Li, and Xiaoyu~Zhang et~al.
\newblock Generative adversarial image super-resolution through deep dense skip
  connections.
\newblock {\em Comput. Graph. Forum}, 37(7):289--300, 2018.

\bibitem[\protect\citeauthoryear{Zou and Yuen}{2012}]{DBLP:journals/tip/ZouY12}
Wilman W.~W. Zou and Pong~C. Yuen.
\newblock Very low resolution face recognition problem.
\newblock {\em {IEEE} Trans. Image Processing}, 21(1):327--340, 2012.

\end{thebibliography}

\end{document}